\documentclass[11pt]{article}
\usepackage[T1]{fontenc}
\usepackage[utf8]{inputenc}
\usepackage{lmodern}
\usepackage{microtype}
\usepackage[affil-it]{authblk}

\usepackage[margin=1in]{geometry}
\usepackage{setspace}
\usepackage{amsmath, amssymb, amsthm}
\usepackage{mathtools}

\usepackage{graphicx}
\usepackage{booktabs}
\usepackage{array}
\usepackage{multirow}
\usepackage{tikz}
\usetikzlibrary{arrows.meta, positioning, shapes.geometric}
\usepackage{xcolor}
\usepackage{enumitem}
\setlist[itemize]{noitemsep, topsep=2pt}
\setlist[enumerate]{noitemsep, topsep=2pt}
\usepackage{hyperref}
\usepackage{cleveref}
\hypersetup{
    colorlinks=true,
    linkcolor=blue!60!black,
    citecolor=green!40!black,
    urlcolor=blue!60!black,
    pdftitle={Data-Centric Anchoring: A Roadmap for Robust and Interpretable Agentic AI},
    pdfauthor={Arun Vignesh Malarkkan, Xinyuan Wang, Yanjie Fu}
}
\usepackage[numbers]{natbib}
\usepackage{mdframed}
\newmdenv[
  backgroundcolor=gray!8,
  linecolor=gray!40,
  linewidth=0.8pt,
  roundcorner=4pt,
  innerleftmargin=10pt,
  innerrightmargin=10pt,
  innertopmargin=8pt,
  innerbottommargin=8pt
]{graybox}
\usepackage{titlesec}
\titleformat{\section}{\large\bfseries}{\thesection.}{0.5em}{}
\titleformat{\subsection}{\normalsize\bfseries}{\thesubsection.}{0.5em}{}
\titleformat{\subsubsection}{\normalsize\itshape}{\thesubsubsection.}{0.5em}{}
\title{Vision: Data-Centric Anchoring for Robust and Interpretable Agentic AI}

\author[]{Arun Vignesh Malarkkan}
\author[]{Xinyuan Wang}
\author[]{Yanjie Fu}
\affil{School of Computing and Augmented Intelligence, Arizona State University, USA}
\affil{\texttt{\{arun.malarkkan, xwang735, yanjie.fu\}@asu.edu}}

\date{}
\begin{document}
\maketitle

\begin{abstract}
Agentic AI systems built on Large Language Models (LLMs) exhibit two persistent failure modes that scaling does not resolve: 1. brittleness under distribution shift, and 2. opacity in decision-making.
We argue that both failures are co-symptoms of a single structural deficiency in the \emph{data lifecycle} governing how agents are trained, evaluated, and deployed.
Observational interaction logs of these agentic systems often encode spurious correlations without controlled variation. As a result, the data lacks the counterfactual structure needed to distinguish causal signals from coincidental patterns or to validate explanations. No model-centric method can recover invariances that the data does not contain.
This paper presents \textbf{Data-Centric Anchoring}, a vision in which robustness and interpretability are engineered into the data environment rather than recovered from models after training. Our primary contribution is the \textbf{Data-Centric Agentic Loop}, a four-stage framework: \textsc{Curate}, \textsc{Augment}, \textsc{Constrain}, and \textsc{Attribute}. The ordering of these stages is not arbitrary, but structurally motivated. Curation must precede augmentation, because generative models tend to amplify the biases present in the data they are trained on. Augmentation must precede constraint enforcement, because invariance objectives require sufficient variation across environments to be meaningful. Finally, attribution closes the loop by converting observed failures into targeted data interventions for the next iteration. This iterative structure makes the framework self-correcting: each stage produces the conditions required for the next.
To ground this framework, we introduce a failure-driven taxonomy that links four core failure modes in the data lifecycle: spurious feature reliance, distribution shift fragility, uncertainty miscalibration, and explanation unfaithfulness. We conclude by identifying the key limitations of this approach and the open challenges that must be resolved to make robust and interpretable agentic AI practical at scale.
\end{abstract}
\section{Introduction}
\label{sec:intro}

Agentic AI systems built on Large Language Models (LLMs) are increasingly deployed in multi-step, interactive settings. Yet their behavior remains unreliable in ways that are difficult to predict,
diagnose, or rectify. They fail under distributional shifts, accumulate errors over long trajectories, and produce explanations that appear coherent but do not reflect the true basis of their decisions. These failures are not rare edge cases but systematic.

At a high level, these issues appear to fall into separate categories. Robustness concerns performance under distribution shift, while interpretability concerns the ability to explain model behavior.
But in practice, the same systems that fail to generalize are also the ones that fail to explain themselves in a verifiable way. This suggests that brittleness and opacity are not distinct failure modes, but symptoms of a shared underlying limitation.
If this limitation were primarily architectural, recent progress in model scaling and training would have mitigated it. Instead, increasingly capable models exhibit the same patterns of failure, often with greater fluency. Despite advances in scaling, alignment, and prompting (e.g., RLHF, chain-of-thought and counterfactual reasoning, and tool-augmented agents) \citep{ouyang2022training, wei2022chain, malarkkan2026finrule, yao2022react}, these methods do not address the structure of the data that determines what can be learned.

Models can only learn invariances that are present in their training distribution, and they can only produce faithful explanations if the data distinguishes causal relationships from coincidental correlations.
Observational interaction logs, which dominate agent training, rarely provide such structure.
They encode correlations without controlled variation, lack counterfactual contrast, and entangle causal and spurious signals. No model-centric intervention can recover invariances or causal annotations
that the data does not contain. This observation reframes the problem. \textbf{Robustness and interpretability are not properties that can be added to a trained model; they must be enabled by the data that shapes it.}
In agentic systems, this data is not static. It evolves across a lifecycle that includes pretraining corpora, synthetic augmentations, interaction logs, and evaluation scenarios. Each stage introduces its own failure modes: spurious correlations during curation, insufficient coverage during augmentation, miscalibration during training, and unfaithful explanations during evaluation.

With this perspective, the reliability gap in agentic AI can be seen as a set of recurring patterns. Agents rely on features that do not generalize beyond the training distribution, fail under shifts in environment or context, express confidence that does not reflect correctness, and produce explanations that cannot be validated under perturbation.
We refer to these as \emph{spurious feature reliance}, \emph{distribution shift fragility}, \emph{uncertainty miscalibration}, and \emph{explanation unfaithfulness}. In agentic settings, these manifest concretely as contextual confounding, trajectory instability, confidence misalignment, and reasoning--decision decoupling. Crucially, each of these failures can be mapped to a deficiency in the data lifecycle. If the problem is rooted in the data lifecycle, then the solution must be as well.
This data-centric perspective motivates our central claim:

\textbf{
Robustness and interpretability in agentic AI are achievable only when the data lifecycle encodes sufficient variation to identify invariant relationships and validate decisions under perturbation; we formalize this
requirement through a data-centric loop that structures how such data is constructed, expanded, and evaluated.
}

\begin{figure}[ht]
    \centering
    \includegraphics[width=\linewidth]{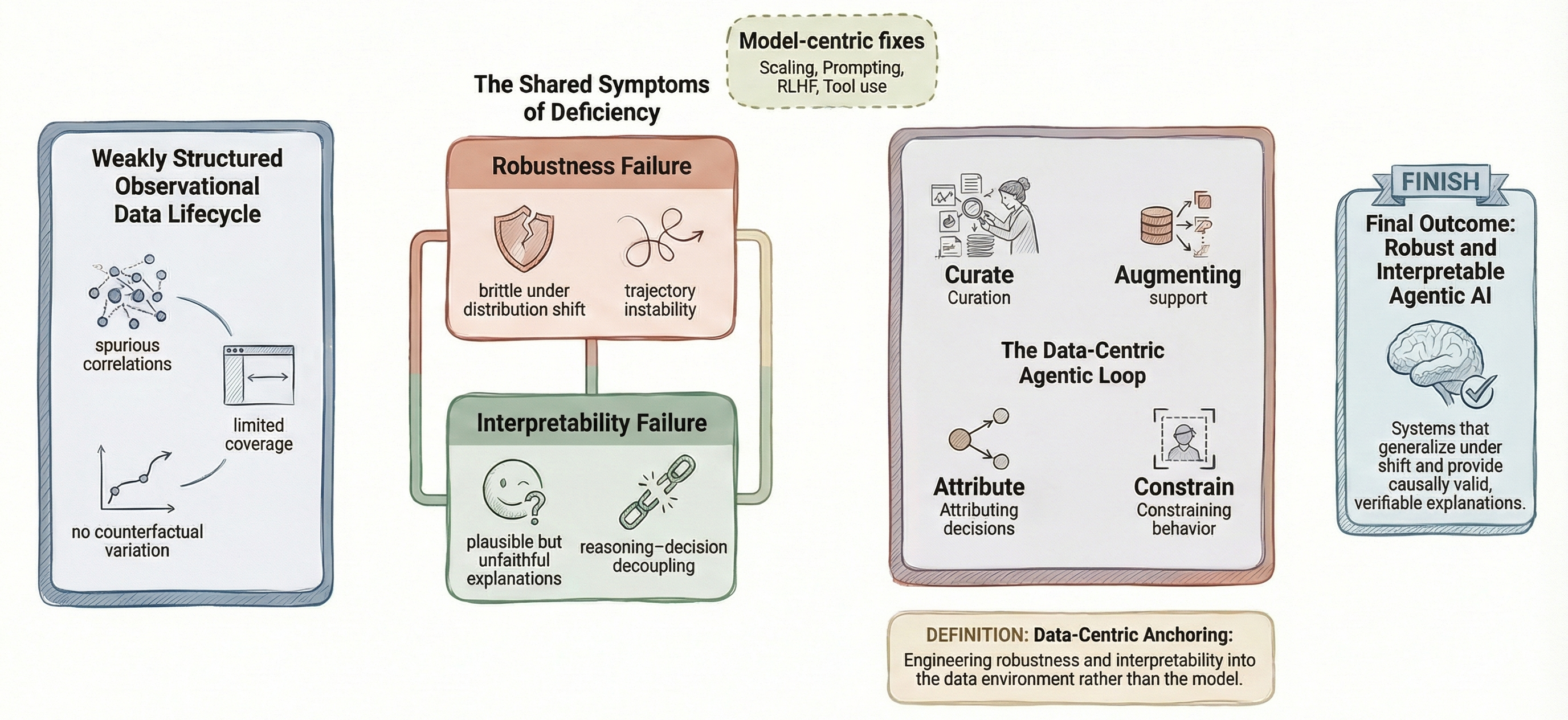}
    \caption{
    Robustness and interpretability failures in agentic AI are not isolated issues, but co-symptoms of a shared structural deficiency in the data lifecycle. Weakly structured observational data lacks the variation needed to support invariant learning and counterfactual validation, motivating a data-centric anchoring perspective.
    }
    \label{fig:motivation}
\end{figure}

This claim has several implications. 
First, robustness and interpretability must be treated as joint objectives. An agent that generalizes without explainability cannot be audited, while an agent that explains without robustness produces unreliable justifications. 
Second, data-centric and model-centric methods are complementary: data defines what invariances are available to learn, while models determine how those invariances are represented and exploited. 
Third, robustness and interpretability cannot be retrofitted after training. They must be enabled during data construction, through sufficient variation to support generalization and validation under perturbation.

As illustrated in Figure~\ref{fig:motivation}, brittleness under distribution shift and explanation unfaithfulness can be understood as two surface manifestations of the same underlying limitation in the data lifecycle.

To operationalize this perspective, we introduce the \textbf{Data-Centric Agentic Loop}, a four-stage framework: \textsc{Curate}, \textsc{Augment}, \textsc{Constrain}, and \textsc{Attribute}.
Curation reduces spurious correlations, augmentation expands coverage across environments, constraint enforcement stabilizes behavior and calibration, and attribution ensures that explanations remain valid under perturbation.

Importantly, each stage can be evaluated through measurable criteria, providing signals for diagnosing failures and guiding subsequent interventions. The stages are not independent: each produces the conditions required for the next, forming an iterative, self-correcting system in which observed failures inform future data refinement. This perspective reframes reliability in agentic AI as a data lifecycle problem, motivating the framework envisioned in the following sections.
\section{A Failure-Driven Taxonomy of Agentic AI}
\label{sec:failures}

The reliability gap in agentic AI systems does not arise from a single deficiency, but from a sequence of structural failures that compound across the data lifecycle. These failures are often treated independently.
However, they are tightly coupled, and addressing one exposes the next. We organize them into four primary modes: spurious feature reliance, distribution shift fragility, uncertainty miscalibration, and explanation unfaithfulness, which together characterize the limits of current systems.

\noindent\underline{\emph{Spurious feature reliance}}: Models trained on observational data learn correlations that are predictive within the training distribution but do not reflect invariant structure. In agentic settings, this appears as contextual confounding, where decisions depend on superficial textual cues such as formatting patterns, phrasing conventions, or latent contextual signals that fail to generalize.
Formally, if $X = (X_c, X_s)$ denotes causal and spurious features, the model may learn $P(Y \mid X_s)$ even when $X_s$ has no effect under intervention.
This behavior is difficult to avoid through model-centric methods: increasing capacity or refining objectives simply strengthens the model’s ability to exploit the strongest available signals in the data.
It can be detected through performance variation across subpopulations, where it appears as \emph{slice leakage}: strong aggregate performance masking systematic errors on specific slices. Mitigation therefore requires modifying the data, like rebalancing slices, introducing counterexamples, or identifying proxy variables, often via automated slice discovery or representation-based clustering.

\noindent\underline{\emph{Distribution shift fragility}}: Correcting spurious feature reliance, however, is not sufficient. Even if a model focuses on more stable features, it may still fail when the input distribution shifts. In agentic systems, this problem is amplified by the sequential nature of interaction, producing cascading failures. This phenomenon, which we refer to as trajectory instability, reflects the fact that models are evaluated on distributions that differ from those seen during training. This gap can be quantified through an \emph{invariance gap}, measuring performance variation across environments or perturbations. Model-centric approaches alone cannot resolve this issue, as invariance cannot be learned without exposure to diverse conditions.
Addressing it requires expanding the support of the data distribution through augmentation, adversarial generation, and environment diversification, closely related to domain randomization and sim-to-real transfer.

\noindent\underline{\emph{Uncertainty miscalibration}}: Robust models may produce predictions with high confidence in regions where they lack sufficient data support. In LLM agents, this appears as confidence misalignment: fluent outputs that do not reflect underlying uncertainty. Standard training objectives optimize likelihood, not calibration, and post-hoc methods only partially correct aggregate discrepancies. Calibration can be evaluated using metrics such as expected calibration error (ECE), but these metrics degrade under distribution shift.
The root cause is again data-related: regions of low density lead to unstable confidence estimates. Improving calibration requires augmenting underrepresented regions and tracking data provenance to understand where uncertainty arises.

\noindent\underline{\emph{Explanation unfaithfulness}}: Finally, even a robust and calibrated system may fail to provide meaningful explanations. In agentic systems, chain-of-thought (CoT) reasoning often appears plausible, but
does not correspond to the true decision mechanism, resulting in reasoning--decision decoupling. Existing interpretability methods typically explain correlations within the model rather than causal dependencies.
As a result, explanations may remain unchanged even when the underlying decision would differ under intervention.
This failure can be detected through \emph{counterfactual faithfulness}: whether modifying explanatory features changes the output. Achieving faithful explanations requires data with sufficient interventional
variation to support such tests. In practice, this is approximated through counterfactual data generation and
attribution methods, though these approaches remain imperfect at scale.

These failures reflect breakdowns at different points in the data lifecycle: curation, coverage, data density, and interventional structure. As a result, robustness and interpretability cannot be achieved through model-centric improvements alone, but require systematic intervention in how data is constructed, expanded, and evaluated.
This observation motivates our vision for a data-centric design principle.

\begin{table}[t]
\centering
\small
\begin{tabular}{p{3.2cm} p{3.2cm} p{4.2cm} p{3.5cm}}
\toprule
\textbf{Failure Mode} & \textbf{Detection Signal} & \textbf{Data-Centric Interventions} & \textbf{Limitations} \\
\midrule

Spurious Feature Reliance &
Slice leakage (performance variation across subgroups) &
Dataset rebalancing, slice discovery, counterexample generation, proxy variable identification &
Requires reliable slice discovery; latent confounders may remain unobserved \\

\midrule

Distribution Shift Fragility &
Invariance gap across environments or perturbations &
Synthetic augmentation, adversarial data generation, environment diversification, domain randomization &
Synthetic data may introduce artifacts; coverage of real-world variation is incomplete \\

\midrule

Uncertainty Miscalibration &
Expected Calibration Error (ECE), confidence–accuracy mismatch &
Augmentation of low-density regions, uncertainty estimation, calibration methods (e.g., temperature scaling) &
Post-hoc calibration does not fix epistemic uncertainty; sensitive to distribution shift \\

\midrule

Explanation Unfaithfulness &
Counterfactual faithfulness (output sensitivity to feature perturbations) &
Data attribution (influence functions, TracIn), provenance tracking, counterfactual testing &
Attribution methods are approximate; counterfactuals depend on valid perturbations \\

\bottomrule
\end{tabular}
\caption{Mapping from failure modes to detection signals, data-centric interventions, and limitations.}
\label{tab:failure_summary}
\end{table}
\section{A Data Lifecycle Perspective on Robustness and Interpretability}
\label{sec:design}

The failure modes identified in \Cref{sec:failures} share a common origin: they arise when the data available to the model does not expose the structure required for robust generalization and reliable explanation. This suggests that reliability is not solely a property of models, but of the data--model system as a whole.
A standard learning formulation treats the dataset as fixed and optimizes model parameters:
\begin{equation}
    \theta^* = \arg\min_{\theta} \mathcal{L}(f_\theta, \mathcal{D})
\end{equation}
This formulation assumes that $\mathcal{D}$ contains sufficient variation to identify invariant relationships and support evaluation. In agentic settings, this assumption is systematically violated. So, we view learning as a joint optimization problem:
\begin{equation}
    (\mathcal{D}^*, \theta^*) = \arg\min_{\mathcal{D}, \theta}
    \mathcal{L}(f_\theta, \mathcal{D})
\end{equation}
where the dataset itself is treated as a design variable. The objective is not only to fit a model, but to construct a data distribution that makes robustness and interpretability attainable.

This reframing shifts the focus from model optimization to system design. Data determines what variation is available, training objectives determine how it is used, and models determine how it is represented. Reliability therefore depends on aligning these components so that the variation exposed by the data supports the invariances the model is expected to learn.
Achieving this requires coordinated data-centric and model-centric
interventions.

We distinguish four types of interventions:

\begin{itemize}
    \item \textbf{Data interventions:} modifying the dataset (e.g., curation,
          augmentation, reweighting),
    \item \textbf{Training interventions:} enforcing objectives (e.g.,
          invariance, calibration),
    \item \textbf{Model interventions:} changing representation or capacity,
    \item \textbf{Evaluation interventions:} diagnosing failures through
          targeted metrics.
\end{itemize}

These interventions are interdependent: data exposes variation, training enforces consistency across it, and models represent the resulting structure. Their effectiveness depends on alignment. Without sufficient variation,
invariance cannot be learned; without coverage, calibration fails; and without perturbations, explanations cannot be validated. This alignment naturally leads to an iterative process, in which failures observed during evaluation guide subsequent data refinement.

We formalize this process as the Data-Centric Agentic Loop.
\section{The Data-Centric Agentic Loop}
\label{sec:loop}

\begin{figure}[ht]
    \centering
    \includegraphics[width=\linewidth]{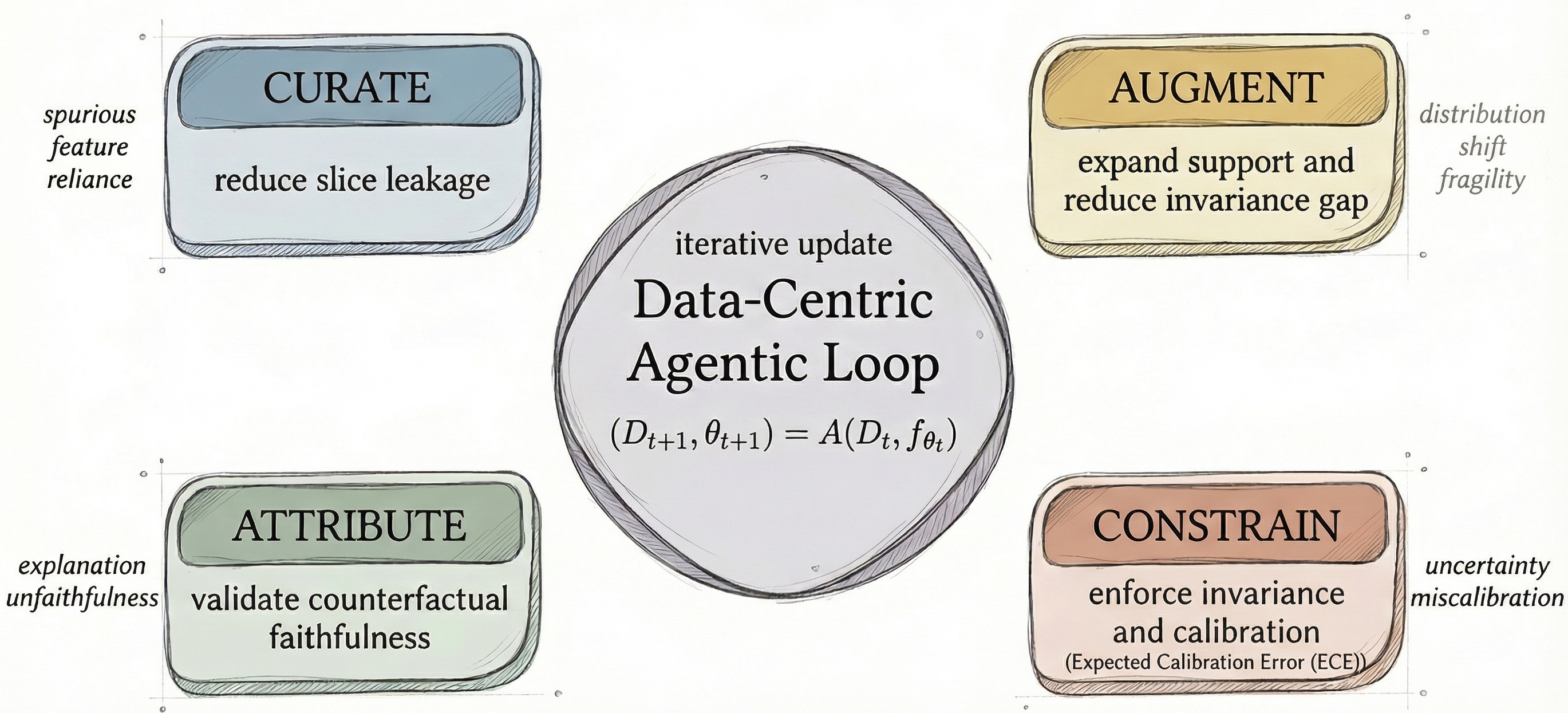}
    \caption{
    The Data-Centric Agentic Loop iteratively refines both the dataset and the agent through four coupled stages: Curate, Augment, Constrain, and Attribute. Each stage addresses a distinct failure signal and produces the conditions required for the next, forming a failure-driven closed loop for robust and interpretable agentic AI.
    }
    \label{fig:loop}
\end{figure}

We formalize the Data-Centric Agentic Loop as an iterative process that jointly updates the data distribution and the agent in order to reduce the failure modes identified in \Cref{sec:failures}.
At iteration $t$, the system consists of a dataset $\mathcal{D}_t$ and an
agent $f_{\theta_t}$, and evolves through updates of the form
\begin{equation}
    (\mathcal{D}_{t+1}, \theta_{t+1}) =
    \mathcal{A}(\mathcal{D}_t, f_{\theta_t}),
\end{equation}
where $\mathcal{A}$ decomposes into four interdependent stages:
\textsc{Curate}, \textsc{Augment}, \textsc{Constrain}, and \textsc{Attribute}.

Figure~\ref{fig:loop} summarizes the proposed Data-Centric Agentic Loop, in which failure signals identified through attribution inform subsequent rounds of curation, augmentation, and constraint enforcement.

\subsection{\textsc{Curate}: Exposing Reliable Signal}

The loop begins with \textsc{Curate}, where the objective is to reduce
spurious feature reliance by restructuring the data distribution.
In observational datasets, predictive performance is often driven by
features that correlate with outcomes but do not reflect invariant
structure.
This manifests empirically as variation in performance across
subpopulations, which we denote as \emph{slice leakage}:
\begin{equation}
    \Delta_{\text{slice}} = \max_{s \in \mathcal{S}} \mathcal{L}_s -
    \min_{s \in \mathcal{S}} \mathcal{L}_s,
\end{equation}
where $\mathcal{S}$ denotes a set of discovered slices.

Curation seeks to reduce $\Delta_{\text{slice}}$ by identifying and
rebalancing these subpopulations.
In practice, slices are discovered through representation-based clustering,
feature probing, or heuristic search over covariates, followed by targeted
data collection or reweighting.
This process weakens the dominance of spurious correlations and exposes
signals that are more stable across contexts.

However, curation is fundamentally limited by observability.
Latent confounders that cannot be detected or approximated through proxies
remain embedded in the data, and errors at this stage propagate forward.
The role of curation is therefore not to eliminate bias completely, but to
make the residual structure more amenable to downstream correction.

\subsection{\textsc{Augment}: Expanding Support}

Even after curation, the dataset remains confined to the support of the
observed distribution.
The \textsc{Augment} stage addresses this by introducing additional variation
through synthetic data, adversarial perturbations, and environment
diversification.

The effectiveness of augmentation can be quantified through an
\emph{invariance gap}, measuring performance variation across environments:
\begin{equation}
    \Delta_{\text{inv}} =
    \max_{e \in \mathcal{E}} \mathcal{L}(f, \mathcal{D}_e) -
    \min_{e \in \mathcal{E}} \mathcal{L}(f, \mathcal{D}_e).
\end{equation}
Reducing $\Delta_{\text{inv}}$ requires exposing the model to sufficient
variation so that invariant features can be identified.

In practice, augmentation introduces new samples that simulate
counterfactual or adversarial conditions.
These may be generated through learned generative models, rule-based
transformations, or simulated environments.
The goal is not merely to increase data volume, but to expand the range of
conditions under which the model is trained.

This stage, however, inherits the limitations of the previous one.
Residual bias in $\mathcal{D}_t$ is propagated and often amplified by
generation, and synthetic environments may introduce artifacts that distort
the data distribution.
Augmentation therefore improves coverage, but does not guarantee correctness.

\subsection{\textsc{Constrain}: Enforcing Consistency}

Given sufficient variation, the \textsc{Constrain} stage enforces consistency
across the expanded data distribution.
Here, the goal is to learn representations and decision rules that are stable
across environments and aligned with uncertainty.

This can be formalized through a worst-case risk objective:
\begin{equation}
    \min_{\theta} \max_{e \in \mathcal{E}} \mathcal{L}(f_\theta, \mathcal{D}_e),
\end{equation}
as well as calibration constraints that align predicted confidence with
empirical accuracy.
In practice, this is implemented through invariant training objectives,
distributionally robust optimization, and calibration techniques such as
temperature scaling or ensembling.

The key dependency is on variation introduced by \textsc{Augment}.
Without sufficient diversity in $\mathcal{D}_t$, invariance cannot be
identified, and constraint-based methods either fail or enforce incorrect
structure.
Thus, constraint enforcement does not create invariance—it filters and
stabilizes what the data already exposes.

\subsection{\textsc{Attribute}: Validating Decisions}

The final stage, \textsc{Attribute}, evaluates whether model decisions can be
explained in terms of the data that produced them.
This is where interpretability is tested, rather than assumed.

We define \emph{counterfactual faithfulness} as the sensitivity of the model's
output to perturbations of explanatory variables:
\begin{equation}
    \Delta_{\text{cf}} =
    \mathbb{E}_{x} \left[ \| f(x) - f(x^{\text{cf}}) \| \right],
\end{equation}
where $x^{\text{cf}}$ denotes a counterfactual modification of $x$ along
explanatory dimensions.

Attribution methods identify candidate explanatory features or training
examples, while counterfactual testing verifies whether these explanations
are causally relevant.
When $\Delta_{\text{cf}}$ is small despite large changes in explanatory
variables, the explanation is unfaithful.

In practice, attribution relies on approximate methods such as influence
functions or gradient tracing, and counterfactual generation depends on
valid perturbations.
As a result, this stage provides noisy but highly informative signals about
model behavior.

\subsection{Iterative Refinement}

The four stages form a coupled system in which each stage transforms the
problem faced by the next.
Curation reduces dominant spurious correlations, enabling more effective
augmentation; augmentation expands variation, enabling meaningful constraint
enforcement; constraints stabilize behavior, enabling reliable attribution;
and attribution identifies failure modes that guide subsequent data
refinement.

This induces a feedback process:
\begin{equation}
    \textsc{Attribute} \rightarrow \textsc{Curate} \rightarrow
    \textsc{Augment} \rightarrow \textsc{Constrain} \rightarrow \textsc{Attribute}.
\end{equation}

While global convergence is not guaranteed, each stage is designed to reduce
a measurable failure signal, and their composition yields a system that
improves through iteration.
Rather than treating robustness and interpretability as properties imposed
on a fixed model, the loop treats them as outcomes of a continuously refined
data--model interaction.
\subsection{Stopping criteria and resource-aware deployment.}
At iteration $k$, let
$m_k = (\Delta_{\text{slice}}^{(k)}, \Delta_{\text{inv}}^{(k)},
\mathrm{ECE}^{(k)}, \Delta_{\text{cf}}^{(k)})$
denote the current diagnostic vector.
We stop iterating when all target metrics fall below user-specified
thresholds, or when the marginal improvement per unit cost drops below a
budget parameter $\lambda$ for $r$ consecutive rounds:
\begin{equation}
    \frac{m_{k-1} - m_k}{\mathrm{cost}_k} < \lambda
    \quad \text{for} \quad r \text{ rounds}.
\end{equation}
This makes the loop budget-aware and prevents overfitting the diagnostics
themselves.

\subsection{Prototype: Operational Instantiation for Language Agents}
\label{ssec:operational-instantiation}

For language-agent deployments, the loop can be instantiated over a set of
trajectories $\{\tau_i\}_{i=1}^n$ and a candidate environment set
$\mathcal{E}_t$ discovered from those trajectories.
We define environments by clustering trajectories along task identity,
tool behavior, interface version, and noise profile:
\begin{equation}
    \mathcal{E}_t
    =
    \mathrm{Cluster}\!\left(
        \{\tau_i\}_{i=1}^n;
        \phi_{\mathrm{task}},
        \phi_{\mathrm{tool}},
        \phi_{\mathrm{interface}},
        \phi_{\mathrm{noise}}
    \right).
\end{equation}

A valid counterfactual $x^{cf}$ is not an arbitrary edit, but a minimal
perturbation that preserves task semantics while remaining execution-valid:
\begin{equation}
    x^{cf}
    \in
    \arg\min_{x' \in \mathcal{P}(x)} d(x,x')
    \quad \text{s.t.} \quad
    \mathrm{Sem}(x') = \mathrm{Sem}(x),
    \;\;
    \mathrm{Exec}(x') \in \mathcal{V},
\end{equation}
where $\mathcal{P}(x)$ denotes a perturbation set, $\mathrm{Sem}(\cdot)$
denotes semantic equivalence, and $\mathcal{V}$ denotes the set of valid
executions.

In practice, $\mathcal{E}_t$ can be approximated using minimal interface
rewrites in the spirit of PIPE~\citep{gu2026agentslearntrajectorysftsemantics} and controlled noise
injection as in AgentNoiseBench~\citep{wang2026agentnoisebenchbenchmarkingrobustnesstoolusing}. A candidate
counterfactual is retained only if it passes executor-based validation and a
semantic-preservation filter.
\section{Limitations}
\label{sec:limitations}

The Data-Centric Agentic Loop assumes that reliability can be improved
through iterative refinement of the data lifecycle.
Its limitations arise when this assumption breaks: when the structure
required is not accessible, not preservable, or not stable.

The first breakdown is observability.
The \textsc{Curate} stage relies on identifying spurious structure through
slice discovery or proxy variables, yet in agentic settings much of the
relevant variation is latent.
Inputs encode implicit user intent, platform conventions, and interaction
history that are only partially observable.
Curation therefore operates on an incomplete view of the data-generating
process: it removes detectable biases while leaving latent confounding
untouched.
What cannot be observed cannot be corrected, and this residual structure
propagates through all subsequent stages.

This incompleteness is amplified, not corrected, by augmentation.
The \textsc{Augment} stage expands the support of the data distribution, but
does so by re-sampling from the structure exposed during curation.
Any residual bias is therefore inherited and often strengthened.
In long-horizon settings, this leads to a deeper failure: synthetic
trajectories remain locally plausible while becoming globally inconsistent.
Augmentation increases variation, but cannot guarantee that the variation is
structurally valid.

Even when the data is expanded, grounding model behavior remains fragile.
The \textsc{Attribute} stage is intended to link decisions to the data that
produced them, but attribution is least reliable precisely where it matters
most.
In low-density regions and at large model scales, influence estimates
degrade, producing explanations that appear grounded but are incorrect.
In such cases, interpretability does not merely fail, but it becomes misleading,
providing confidence without justification.

Causal reasoning introduces a more subtle failure.
Unlike empirical methods, which degrade visibly, causal methods fail
silently under misspecification.
An incorrect structural assumption does not simply reduce performance; it
propagates through the loop, shaping augmentation, and validating
explanations against an incorrect criterion.
Because these errors do not necessarily surface in standard metrics, they
can remain undetected while reinforcing incorrect structure.

These limitations converge in the most fundamental constraint: the loop
assumes a stable target distribution.
In agentic systems, this assumption does not hold.
The data distribution is shaped by the agent itself,
\begin{equation}
    \mathcal{D}_{t+1} = \mathcal{F}(\mathcal{D}_t, f_{\theta_t}),
\end{equation}
so that each iteration modifies the environment it is meant to model.
This performative non-stationarity breaks the premise of iterative
refinement.
The system is not converging toward a fixed distribution; it is continually
reshaping it.

Taken together, these limitations reveal a deeper constraint:
the effectiveness of the loop is bounded not by any single stage, but by the
extent to which the data lifecycle itself can be observed, controlled, and
stabilized.

\section{Future Directions}
\label{sec:challenges}

The limitations above define a corresponding research agenda.
Each direction addresses a specific structural constraint in the framework,
rather than extending it in isolation.

Addressing observability requires scalable methods for discovering and
representing latent structure in high-dimensional data.
This includes advances in automated slice discovery, representation probing,
and methods that incorporate uncertainty over unobserved variables rather
than relying on point estimates.
Progress in this direction would expand the effectiveness of the
\textsc{Curate} stage beyond observable correlations.

Improving augmentation requires moving from distributional coverage to
structural validity.
Future work must ensure that synthetic data preserves the causal and temporal
consistency of the underlying process, particularly for long-horizon
trajectories.
This includes developing metrics that detect structural inconsistencies and
methods that generate data under multiple plausible environments rather than
a single learned distribution.

Reliable attribution requires new paradigms for grounding decisions in data.
Scalable alternatives to gradient-based influence estimation, as well as
methods for compositional attribution across multi-step trajectories, are
necessary.
Equally important is the development of evaluation protocols that detect when
explanations are misleading rather than merely incomplete.

Robust integration of causal reasoning requires methods that tolerate
uncertain or partially incorrect structural assumptions.
This includes incorporating sensitivity analysis, uncertainty over causal
graphs, and hybrid approaches that combine causal reasoning with empirical
validation rather than treating causal structure as fixed.

Finally, addressing non-stationarity requires rethinking the loop as a
dynamic system rather than a static optimization process.
Future work must account for the fact that the agent co-produces its data
distribution, requiring mechanisms that track the effects of past
interventions, update structural assumptions over time, and anticipate
future distribution shifts rather than reacting to them.

Together, these directions point toward a broader shift: from improving
individual components of the learning pipeline to designing systems that
explicitly account for the interaction between data, models, and the
environments they shape.

\paragraph{Evaluation Protocol.}
A natural next step is a small-scale synthetic evaluation environment with
known environment labels, controllable interface perturbations, and known
counterfactual ground truth.
Such a protocol would support ablations that remove \textsc{Curate},
\textsc{Augment}, \textsc{Constrain}, or \textsc{Attribute} in isolation,
allowing the loop to be evaluated using
$\Delta_{\text{slice}}$, $\Delta_{\text{inv}}$, $\mathrm{ECE}$, and
$\Delta_{\text{cf}}$ without claiming a full empirical validation.
\section{Related Work}
\label{sec:related_work}

\subsection{Data-Centric AI and the Data Lifecycle}
A central premise of this paper is that robustness and interpretability are not solely model properties, but emerge from the structure of the data lifecycle. This view builds on data-centric AI, which shifts attention from model design to training-data quality, inference-data quality, and ongoing data maintenance \citep{zha2023data,zha2025data, ying2025survey, malarkkan2025delta}. More broadly, this direction is consistent with earlier work on data valuation, dataset debugging, and data attribution, which seeks to quantify the effect of individual examples or subsets on downstream model behavior
\citep{ghorbani2019data,jia2019towards,koh2017understanding,yeh2018representer,pruthi2020estimating,hammoudeh2024training}. Recent tutorials and surveys have further clarified that data-centric methods span curation, augmentation, maintenance, and evaluation, rather than any single intervention point \citep{zha2023data,hu2025snapshot}. Our framing differs from these works in that we treat the entire data lifecycle as a coupled system whose failures are organized around robustness and interpretability rather than data quality in the abstract.

\subsection{Robustness, Domain Generalization, and Invariance}
The robustness side of our framework is closely related to domain generalization, distributionally robust optimization, and invariance learning. Domain generalization surveys summarize a broad family of approaches that seek out-of-distribution generalization under domain shift, including data augmentation, meta-learning, ensemble methods, and representation learning
\citep{zhou2022domain,khoee2024domain,rafi2024domain,zhong2025survey}. Within this literature, IRM formalized the goal of learning representations whose optimal classifier is stable across environments, and group DRO and related formulations optimize worst-group or worst-environment risk \citep{arjovsky2019invariant,sagawa2019distributionally,krueger2021out,rahimian2019distributionally,mehta2023distributionally}. These ideas are also connected to sim-to-real transfer and domain randomization, where the training distribution is deliberately broadened to expose invariances that survive environmental change \citep{tobin2017domain,peng2018sim}. Our perspective differs in two ways. First, we treat invariance not as a standalone objective but as one stage in a larger data-centric loop. Second, we explicitly connect invariance to interpretability, since the same data variations that expose stable structure are also needed to validate
counterfactual explanations.

\subsection{Interpretability, Data Attribution, and Faithful Explanations}
Classic influence-function methods trace model predictions back to training examples, and representer-point methods provide a scalable alternative for linking predictions to influential data points \citep{koh2017understanding,yeh2018representer}. Recent work on training-data influence analysis and data attribution has expanded this line of research, including scalable approximations and broader taxonomies of attribution goals \citep{pruthi2020estimating, hammoudeh2024training,ratul2021evaluating}. At the explanation level, however, a major challenge is faithfulness. Chain-of-thought prompts can improve performance, but multiple studies show that reasoning traces are not necessarily faithful explanations of the underlying computation \citep{lyu2023faithful,lanham2023measuring, paul2024making}. This motivates our emphasis on counterfactual validation rather than surface-level explanation. Our goal is not merely to produce plausible rationales, but to check whether changing the factors named in an explanation actually changes the decision. That criterion is closer to causal validity than to standard feature attribution, and it is the reason the \textsc{Attribute} stage in our loop is defined around perturbation and provenance.

\subsection{Agentic Evaluation, Benchmarks, and Causal Grounding}
Recent surveys show that agent evaluation has matured into a distinct research area, with benchmarks spanning planning, tool use, memory, robustness, and cost-efficiency \citep{yehudai2025survey}. Representative benchmarks and evaluation frameworks include AgentBoard, AgentQuest, GTA, Mobile-Bench, SmartPlay, and MixEval, each of which highlights gaps between toy task success and realistic multi-step behavior \citep{ma2024agentboard,gioacchini2024agentquest,wang2024gta,deng2024mobile,wu2023smartplay,ni2024mixeval}.
These efforts motivate our claim that agentic systems should be evaluated through failure signals that are visible across stages, not only by final task success. A related but distinct line of work uses causal reasoning for discovery, counterfactual generation, and explanation. Surveys on causal inference with LLMs, causal generative modeling, and causality more broadly show growing interest in using causal language and interventional reasoning to support learning and explanation \citep{ma2025causal,komanduri2023identifiable, lamsaf2025causality, 11314741, malarkkan2026causallyguidedautomatedfeatureengineering,malarkkan2025rethinkingspatiotemporalanomalydetection}.
We draw on this literature selectively: causality is useful when the assumptions behind interventional reasoning are approximately satisfied, but our framework does not require every stage to be causal in the formal sense. Instead, causality is one tool within a broader data-centric roadmap that also includes curation, augmentation, calibration, and attribution.

\subsection{Agentic Robustness Benchmarks and Data Debugging}
Recent benchmark work suggests that agent failures should be diagnosed rather than only scored. PIPE~\cite{gu2026agentslearntrajectorysftsemantics} exposes interface shortcutting by minimally rewriting environment interfaces while preserving semantics and execution behavior; AgentNoiseBench~\cite{wang2026agentnoisebenchbenchmarkingrobustnesstoolusing} injects controllable user-noise and tool-noise into existing benchmarks; and PALADIN~\cite{vuddanti2025paladinselfcorrectinglanguagemodel} trains on recovery-annotated trajectories constructed via systematic failure injection and expert demonstrations. These works align naturally with the \textsc{Augment} and \textsc{Attribute} stages of our loop. On the data side, Snorkel~\cite{merrill2026terminalbenchbenchmarkingagentshard} operationalizes weak supervision and data programming, while Cleanlab~\cite{northcutt2022confidentlearningestimatinguncertainty} implements confident learning for noisy-label detection. For scalable attribution, TRAK~\cite{park2023trakattributingmodelbehavior} and Daunce~\cite{pan2025dauncedataattributionuncertainty} provide practical alternatives to classical influence estimation.
Existing agent evaluation suites such as AgentBoard~\cite{ma2024agentboardanalyticalevaluationboard}, GTA~\cite{wang2024gtabenchmarkgeneraltool}, SmartPlay~\cite{wu2024smartplaybenchmarkllmsintelligent}, and MixEval~\cite{ni2024mixevalderivingwisdomcrowd} already cover general multi-turn and tool-use behavior, but they do not directly expose slice leakage, invariance gaps, or counterfactual faithfulness.

Across these literatures, existing work addresses important pieces of the problem: data quality, invariance, calibration, attribution, and agent evaluation. Our contribution is to connect them through a single failure-driven loop that treats the data lifecycle as the primary object of design for robust and interpretable agentic AI.

\begin{table}[t]
\centering
\small
\begin{tabular}{p{2.1cm} p{3.2cm} p{3.7cm} p{3.0cm}}
\toprule
\textbf{Loop stage} & \textbf{Relevant benchmarks / tools} & \textbf{What they diagnose} & \textbf{Suggested metric} \\
\midrule
Curate &
Cleanlab, Snorkel, PIPE &
Noisy labels, weak supervision, interface shortcutting, slice leakage &
$\Delta_{\text{slice}}$ \\
\midrule
Augment &
AgentNoiseBench, counterfactual augmentation, PALADIN &
Robustness under user-noise, tool-noise, and failure injection &
$\Delta_{\text{inv}}$ \\
\midrule
Constrain &
Group DRO, IRM, ASGDRO, calibration methods &
Worst-group instability, invariance failure, miscalibration &
Worst-group risk, ECE \\
\midrule
Attribute &
TRAK, Daunce, influence functions, representer points &
Training-data influence and explanation faithfulness &
$\Delta_{\text{cf}}$ \\
\bottomrule
\end{tabular}
\caption{A concrete mapping from loop stages to benchmarks, debugging tools, and diagnostics.}
\label{tab:loop_benchmark_map}
\end{table}
\section{Conclusion}
\label{sec:conclusion}

Agentic AI systems fail not because models are insufficiently powerful, but because the data they are trained on does not encode the variation required for reliable generalization and verifiable explanation. Brittleness under distribution shift and opacity in reasoning are therefore not separate problems, but consequences of the same constraint: invariant structure cannot be learned or tested without appropriate variation in the data. This perspective shifts the focus of progress. Improving model architectures or training objectives cannot compensate for missing structure in the data. Reliability becomes a question of whether the data lifecycle exposes the
right variation, at the right granularity, and under the right conditions. Viewed in this light, robustness and interpretability are not properties to be added to trained systems, but properties that must be enabled by design.
The Data-Centric Agentic Loop provides one abstraction for this process, but its significance lies not in its specific stages, but in the shift it represents: from optimizing models over fixed data to engineering the data
itself as part of the learning system. The path forward is therefore not only to build more capable models, but to
construct data environments in which reliable and verifiable behavior is possible.
\newpage
\bibliographystyle{plainnat}
\bibliography{main}

@article{ouyang2022training,
  title={Training language models to follow instructions with human feedback},
  author={Ouyang, Long and Wu, Jeffrey and Jiang, Xu and Almeida, Diogo and Wainwright, Carroll and Mishkin, Pamela and Zhang, Chong and Agarwal, Sandhini and Slama, Katarina and Ray, Alex and others},
  journal={Advances in neural information processing systems},
  volume={35},
  pages={27730--27744},
  year={2022}
}

@article{wei2022chain,
  title={Chain-of-thought prompting elicits reasoning in large language models},
  author={Wei, Jason and Wang, Xuezhi and Schuurmans, Dale and Bosma, Maarten and Xia, Fei and Chi, Ed and Le, Quoc V and Zhou, Denny and others},
  journal={Advances in neural information processing systems},
  volume={35},
  pages={24824--24837},
  year={2022}
}

@article{malarkkan2026finrule,
  title={FinRule-Bench: A Benchmark for Joint Reasoning over Financial Tables and Principles},
  author={Malarkkan, Arun Vignesh and Choudhury, Manan Roy and Zhang, Guangwei and Gupta, Vivek and Wang, Qingyun and Fu, Yanjie and Zhang, Denghui},
  journal={arXiv preprint arXiv:2603.11339},
  year={2026}
}

@inproceedings{yao2022react,
  title={React: Synergizing reasoning and acting in language models},
  author={Yao, Shunyu and Zhao, Jeffrey and Yu, Dian and Du, Nan and Shafran, Izhak and Narasimhan, Karthik R and Cao, Yuan},
  booktitle={The eleventh international conference on learning representations},
  year={2022}
}

@article{zha2025data,
  title={Data-centric artificial intelligence: A survey},
  author={Zha, Daochen and Bhat, Zaid Pervaiz and Lai, Kwei-Herng and Yang, Fan and Jiang, Zhimeng and Zhong, Shaochen and Hu, Xia},
  journal={ACM Computing Surveys},
  volume={57},
  number={5},
  pages={1--42},
  year={2025},
  publisher={ACM New York, NY}
}

@inproceedings{zha2023data,
  title={Data-centric ai: Perspectives and challenges},
  author={Zha, Daochen and Bhat, Zaid Pervaiz and Lai, Kwei-Herng and Yang, Fan and Hu, Xia},
  booktitle={Proceedings of the 2023 SIAM international conference on data mining (SDM)},
  pages={945--948},
  year={2023},
  organization={SIAM}
}

@article{ying2025survey,
  title={A survey on data-centric ai: Tabular learning from reinforcement learning and generative ai perspective},
  author={Ying, Wangyang and Wei, Cong and Gong, Nanxu and Wang, Xinyuan and Bai, Haoyue and Malarkkan, Arun Vignesh and Dong, Sixun and Wang, Dongjie and Zhang, Denghui and Fu, Yanjie},
  journal={arXiv preprint arXiv:2502.08828},
  year={2025}
}

@inproceedings{ghorbani2019data,
  title={Data shapley: Equitable valuation of data for machine learning},
  author={Ghorbani, Amirata and Zou, James},
  booktitle={International conference on machine learning},
  pages={2242--2251},
  year={2019},
  organization={PMLR}
}

@inproceedings{jia2019towards,
  title={Towards efficient data valuation based on the shapley value},
  author={Jia, Ruoxi and Dao, David and Wang, Boxin and Hubis, Frances Ann and Hynes, Nick and G{\"u}rel, Nezihe Merve and Li, Bo and Zhang, Ce and Song, Dawn and Spanos, Costas J},
  booktitle={The 22nd international conference on artificial intelligence and statistics},
  pages={1167--1176},
  year={2019},
  organization={PMLR}
}

@inproceedings{koh2017understanding,
  title={Understanding black-box predictions via influence functions},
  author={Koh, Pang Wei and Liang, Percy},
  booktitle={International conference on machine learning},
  pages={1885--1894},
  year={2017},
  organization={PMLR}
}

@article{yeh2018representer,
  title={Representer point selection for explaining deep neural networks},
  author={Yeh, Chih-Kuan and Kim, Joon and Yen, Ian En-Hsu and Ravikumar, Pradeep K},
  journal={Advances in neural information processing systems},
  volume={31},
  year={2018}
}

@article{pruthi2020estimating,
  title={Estimating training data influence by tracing gradient descent},
  author={Pruthi, Garima and Liu, Frederick and Kale, Satyen and Sundararajan, Mukund},
  journal={Advances in Neural Information Processing Systems},
  volume={33},
  pages={19920--19930},
  year={2020}
}

@article{hammoudeh2024training,
  title={Training data influence analysis and estimation: A survey},
  author={Hammoudeh, Zayd and Lowd, Daniel},
  journal={Machine Learning},
  volume={113},
  number={5},
  pages={2351--2403},
  year={2024},
  publisher={Springer}
}

@article{hu2025snapshot,
  title={A snapshot of influence: A local data attribution framework for online reinforcement learning},
  author={Hu, Yuzheng and Wu, Fan and Ye, Haotian and Forsyth, David and Zou, James and Jiang, Nan and Ma, Jiaqi W and Zhao, Han},
  journal={arXiv preprint arXiv:2505.19281},
  year={2025}
}

@article{zhou2022domain,
  title={Domain generalization: A survey},
  author={Zhou, Kaiyang and Liu, Ziwei and Qiao, Yu and Xiang, Tao and Loy, Chen Change},
  journal={IEEE transactions on pattern analysis and machine intelligence},
  volume={45},
  number={4},
  pages={4396--4415},
  year={2022},
  publisher={IEEE}
}

@article{khoee2024domain,
  title={Domain generalization through meta-learning: a survey},
  author={Khoee, Arsham Gholamzadeh and Yu, Yinan and Feldt, Robert},
  journal={Artificial Intelligence Review},
  volume={57},
  number={10},
  pages={285},
  year={2024},
  publisher={Springer}
}

@article{rafi2024domain,
  title={Domain generalization for semantic segmentation: a survey},
  author={Rafi, Taki Hasan and Mahjabin, Ratul and Ghosh, Emon and Ko, Young-Woong and Lee, Jeong-Gun},
  journal={Artificial Intelligence Review},
  volume={57},
  number={9},
  pages={247},
  year={2024},
  publisher={Springer}
}

@article{zhong2025survey,
  title={A survey of data augmentation in domain generalization},
  author={Zhong, Yingyi and Zhou, Wen’an and Wang, Zhixian},
  journal={Neural Processing Letters},
  volume={57},
  number={2},
  pages={34},
  year={2025},
  publisher={Springer}
}

@article{arjovsky2019invariant,
  title={Invariant risk minimization},
  author={Arjovsky, Martin and Bottou, L{\'e}on and Gulrajani, Ishaan and Lopez-Paz, David},
  journal={arXiv preprint arXiv:1907.02893},
  year={2019}
}

@article{sagawa2019distributionally,
  title={Distributionally robust neural networks for group shifts: On the importance of regularization for worst-case generalization},
  author={Sagawa, Shiori and Koh, Pang Wei and Hashimoto, Tatsunori B and Liang, Percy},
  journal={arXiv preprint arXiv:1911.08731},
  year={2019}
}

@inproceedings{krueger2021out,
  title={Out-of-distribution generalization via risk extrapolation (rex)},
  author={Krueger, David and Caballero, Ethan and Jacobsen, Joern-Henrik and Zhang, Amy and Binas, Jonathan and Zhang, Dinghuai and Le Priol, Remi and Courville, Aaron},
  booktitle={International conference on machine learning},
  pages={5815--5826},
  year={2021},
  organization={PMLR}
}

@article{rahimian2019distributionally,
  title={Distributionally robust optimization: A review},
  author={Rahimian, Hamed and Mehrotra, Sanjay},
  journal={arXiv preprint arXiv:1908.05659},
  year={2019}
}

@article{mehta2023distributionally,
  title={Distributionally robust optimization with bias and variance reduction},
  author={Mehta, Ronak and Roulet, Vincent and Pillutla, Krishna and Harchaoui, Zaid},
  journal={arXiv preprint arXiv:2310.13863},
  year={2023}
}

@inproceedings{tobin2017domain,
  title={Domain randomization for transferring deep neural networks from simulation to the real world},
  author={Tobin, Josh and Fong, Rachel and Ray, Alex and Schneider, Jonas and Zaremba, Wojciech and Abbeel, Pieter},
  booktitle={2017 IEEE/RSJ international conference on intelligent robots and systems (IROS)},
  pages={23--30},
  year={2017},
  organization={IEEE}
}

@inproceedings{peng2018sim,
  title={Sim-to-real transfer of robotic control with dynamics randomization},
  author={Peng, Xue Bin and Andrychowicz, Marcin and Zaremba, Wojciech and Abbeel, Pieter},
  booktitle={2018 IEEE international conference on robotics and automation (ICRA)},
  pages={3803--3810},
  year={2018},
  organization={IEEE}
}

@inproceedings{ratul2021evaluating,
  title={Evaluating attribution methods in machine learning interpretability},
  author={Ratul, Qudrat E Alahy and Serra, Edoardo and Cuzzocrea, Alfredo},
  booktitle={2021 IEEE International Conference on Big Data (Big Data)},
  pages={5239--5245},
  year={2021},
  organization={IEEE}
}

@inproceedings{lyu2023faithful,
  title={Faithful chain-of-thought reasoning},
  author={Lyu, Qing and Havaldar, Shreya and Stein, Adam and Zhang, Li and Rao, Delip and Wong, Eric and Apidianaki, Marianna and Callison-Burch, Chris},
  booktitle={Proceedings of the 13th International Joint Conference on Natural Language Processing and the 3rd Conference of the Asia-Pacific Chapter of the Association for Computational Linguistics (Volume 1: Long Papers)},
  pages={305--329},
  year={2023}
}

@article{lanham2023measuring,
  title={Measuring faithfulness in chain-of-thought reasoning},
  author={Lanham, Tamera and Chen, Anna and Radhakrishnan, Ansh and Steiner, Benoit and Denison, Carson and Hernandez, Danny and Li, Dustin and Durmus, Esin and Hubinger, Evan and Kernion, Jackson and others},
  journal={arXiv preprint arXiv:2307.13702},
  year={2023}
}

@inproceedings{paul2024making,
  title={Making reasoning matter: Measuring and improving faithfulness of chain-of-thought reasoning},
  author={Paul, Debjit and West, Robert and Bosselut, Antoine and Faltings, Boi},
  booktitle={Findings of the Association for Computational Linguistics: EMNLP 2024},
  pages={15012--15032},
  year={2024}
}

@article{yehudai2025survey,
  title={Survey on evaluation of llm-based agents},
  author={Yehudai, Asaf and Eden, Lilach and Li, Alan and Uziel, Guy and Zhao, Yilun and Bar-Haim, Roy and Cohan, Arman and Shmueli-Scheuer, Michal},
  journal={arXiv preprint arXiv:2503.16416},
  year={2025}
}

@article{ma2024agentboard,
  title={Agentboard: An analytical evaluation board of multi-turn llm agents},
  author={Ma, Chang and Zhang, Junlei and Zhu, Zhihao and Yang, Cheng and Yang, Yujiu and Jin, Yaohui and Lan, Zhenzhong and Kong, Lingpeng and He, Junxian},
  journal={Advances in neural information processing systems},
  volume={37},
  pages={74325--74362},
  year={2024}
}

@inproceedings{gioacchini2024agentquest,
  title={Agentquest: A modular benchmark framework to measure progress and improve llm agents},
  author={Gioacchini, Luca and Siracusano, Giuseppe and Sanvito, Davide and Gashteovski, Kiril and Friede, David and Bifulco, Roberto and Lawrence, Carolin},
  booktitle={Proceedings of the 2024 Conference of the North American Chapter of the Association for Computational Linguistics: Human Language Technologies (Volume 3: System Demonstrations)},
  pages={185--193},
  year={2024}
}

@article{wang2024gta,
  title={GTA: a benchmark for general tool agents},
  author={Wang, Jize and Ma, Zerun and Li, Yining and Zhang, Songyang and Chen, Cailian and Chen, Kai and Le, Xinyi},
  journal={Advances in Neural Information Processing Systems},
  volume={37},
  pages={75749--75790},
  year={2024}
}

@inproceedings{deng2024mobile,
  title={Mobile-bench: An evaluation benchmark for llm-based mobile agents},
  author={Deng, Shihan and Xu, Weikai and Sun, Hongda and Liu, Wei and Tan, Tao and Liujianfeng, Liujianfeng and Li, Ang and Luan, Jian and Wang, Bin and Yan, Rui and others},
  booktitle={Proceedings of the 62nd Annual Meeting of the Association for Computational Linguistics (Volume 1: Long Papers)},
  pages={8813--8831},
  year={2024}
}

@article{wu2023smartplay,
  title={Smartplay: A benchmark for llms as intelligent agents},
  author={Wu, Yue and Tang, Xuan and Mitchell, Tom M and Li, Yuanzhi},
  journal={arXiv preprint arXiv:2310.01557},
  year={2023}
}

@article{ni2024mixeval,
  title={Mixeval: Deriving wisdom of the crowd from llm benchmark mixtures},
  author={Ni, Jinjie and Xue, Fuzhao and Yue, Xiang and Deng, Yuntian and Shah, Mahir and Jain, Kabir and Neubig, Graham and You, Yang},
  journal={Advances in Neural Information Processing Systems},
  volume={37},
  pages={98180--98212},
  year={2024}
}

@article{ma2025causal,
  title={Causal inference with large language model: A survey},
  author={Ma, Jing},
  journal={Findings of the Association for Computational Linguistics: NAACL 2025},
  pages={5886--5898},
  year={2025}
}

@article{komanduri2023identifiable,
  title={From identifiable causal representations to controllable counterfactual generation: A survey on causal generative modeling},
  author={Komanduri, Aneesh and Wu, Xintao and Wu, Yongkai and Chen, Feng},
  journal={arXiv preprint arXiv:2310.11011},
  year={2023}
}

@article{lamsaf2025causality,
  title={Causality, machine learning, and feature selection: a survey},
  author={Lamsaf, Asmae and Carrilho, Rui and Neves, Jo{\~a}o C and Proen{\c{c}}a, Hugo},
  journal={Sensors},
  volume={25},
  number={8},
  pages={2373},
  year={2025},
  publisher={MDPI}
}

@article{malarkkan2025delta,
  title={DELTA: Variational Disentangled Learning for Privacy-Preserving Data Reprogramming},
  author={Malarkkan, Arun Vignesh and Bai, Haoyue and Kaushik, Anjali and Fu, Yanjie},
  journal={arXiv preprint arXiv:2509.00693},
  year={2025}
}

@ARTICLE{11314741,
  author={Malarkkan, Arun Vignesh and Wang, Dongjie and Bai, Haoyue and Fu, Yanjie},
  journal={IEEE Transactions on Big Data}, 
  title={Incremental Causal Graph Learning for Online Cyberattack Detection in Cyber-Physical Infrastructures}, 
  year={2025},
  volume={},
  number={},
  pages={1-12},
  doi={10.1109/TBDATA.2025.3648314}}

@misc{malarkkan2026causallyguidedautomatedfeatureengineering,
      title={Causally-Guided Automated Feature Engineering with Multi-Agent Reinforcement Learning}, 
      author={Arun Vignesh Malarkkan and Wangyang Ying and Yanjie Fu},
      year={2026},
      eprint={2602.16435},
      archivePrefix={arXiv},
      primaryClass={cs.AI},
      url={https://arxiv.org/abs/2602.16435}, 
}

@misc{malarkkan2025rethinkingspatiotemporalanomalydetection,
      title={Rethinking Spatio-Temporal Anomaly Detection: A Vision for Causality-Driven Cybersecurity}, 
      author={Arun Vignesh Malarkkan and Haoyue Bai and Xinyuan Wang and Anjali Kaushik and Dongjie Wang and Yanjie Fu},
      year={2025},
      eprint={2507.08177},
      archivePrefix={arXiv},
      primaryClass={cs.LG},
      url={https://arxiv.org/abs/2507.08177}, 
}

@misc{gu2026agentslearntrajectorysftsemantics,
      title={What Do Agents Learn from Trajectory-SFT: Semantics or Interfaces?}, 
      author={Weizheng Gu and Chengze Li and Zhuohao Yu and Mengyuan Sun and Zhibang Yang and Wei Wang and Hongrui Jia and Shikun Zhang and Wei Ye},
      year={2026},
      eprint={2602.01611},
      archivePrefix={arXiv},
      primaryClass={cs.LG},
      url={https://arxiv.org/abs/2602.01611}, 
}

@misc{wang2026agentnoisebenchbenchmarkingrobustnesstoolusing,
      title={AgentNoiseBench: Benchmarking Robustness of Tool-Using LLM Agents Under Noisy Condition}, 
      author={Ruipeng Wang and Yuxin Chen and Yukai Wang and Chang Wu and Junfeng Fang and Xiaodong Cai and Qi Gu and Hui Su and An Zhang and Xiang Wang and Xunliang Cai and Tat-Seng Chua},
      year={2026},
      eprint={2602.11348},
      archivePrefix={arXiv},
      primaryClass={cs.AI},
      url={https://arxiv.org/abs/2602.11348}, 
}

@misc{vuddanti2025paladinselfcorrectinglanguagemodel,
      title={PALADIN: Self-Correcting Language Model Agents to Cure Tool-Failure Cases}, 
      author={Sri Vatsa Vuddanti and Aarav Shah and Satwik Kumar Chittiprolu and Tony Song and Sunishchal Dev and Kevin Zhu and Maheep Chaudhary},
      year={2025},
      eprint={2509.25238},
      archivePrefix={arXiv},
      primaryClass={cs.LG},
      url={https://arxiv.org/abs/2509.25238}, 
}

@misc{merrill2026terminalbenchbenchmarkingagentshard,
      title={Terminal-Bench: Benchmarking Agents on Hard, Realistic Tasks in Command Line Interfaces}, 
      author={Mike A. Merrill and Alexander G. Shaw and Nicholas Carlini and Boxuan Li and Harsh Raj et al.},
      year={2026},
      eprint={2601.11868},
      archivePrefix={arXiv},
      primaryClass={cs.SE},
      url={https://arxiv.org/abs/2601.11868}, 
}

@misc{northcutt2022confidentlearningestimatinguncertainty,
      title={Confident Learning: Estimating Uncertainty in Dataset Labels}, 
      author={Curtis G. Northcutt and Lu Jiang and Isaac L. Chuang},
      year={2022},
      eprint={1911.00068},
      archivePrefix={arXiv},
      primaryClass={stat.ML},
      url={https://arxiv.org/abs/1911.00068}, 
}

@misc{park2023trakattributingmodelbehavior,
      title={TRAK: Attributing Model Behavior at Scale}, 
      author={Sung Min Park and Kristian Georgiev and Andrew Ilyas and Guillaume Leclerc and Aleksander Madry},
      year={2023},
      eprint={2303.14186},
      archivePrefix={arXiv},
      primaryClass={stat.ML},
      url={https://arxiv.org/abs/2303.14186}, 
}

@misc{pan2025dauncedataattributionuncertainty,
      title={Daunce: Data Attribution through Uncertainty Estimation}, 
      author={Xingyuan Pan and Chenlu Ye and Joseph Melkonian and Jiaqi W. Ma and Tong Zhang},
      year={2025},
      eprint={2505.23223},
      archivePrefix={arXiv},
      primaryClass={cs.LG},
      url={https://arxiv.org/abs/2505.23223}, 
}

@misc{ma2024agentboardanalyticalevaluationboard,
      title={AgentBoard: An Analytical Evaluation Board of Multi-turn LLM Agents}, 
      author={Chang Ma and Junlei Zhang and Zhihao Zhu and Cheng Yang and Yujiu Yang and Yaohui Jin and Zhenzhong Lan and Lingpeng Kong and Junxian He},
      year={2024},
      eprint={2401.13178},
      archivePrefix={arXiv},
      primaryClass={cs.CL},
      url={https://arxiv.org/abs/2401.13178}, 
}

@misc{wang2024gtabenchmarkgeneraltool,
      title={GTA: A Benchmark for General Tool Agents}, 
      author={Jize Wang and Zerun Ma and Yining Li and Songyang Zhang and Cailian Chen and Kai Chen and Xinyi Le},
      year={2024},
      eprint={2407.08713},
      archivePrefix={arXiv},
      primaryClass={cs.CL},
      url={https://arxiv.org/abs/2407.08713}, 
}

@misc{wu2024smartplaybenchmarkllmsintelligent,
      title={SmartPlay: A Benchmark for LLMs as Intelligent Agents}, 
      author={Yue Wu and Xuan Tang and Tom M. Mitchell and Yuanzhi Li},
      year={2024},
      eprint={2310.01557},
      archivePrefix={arXiv},
      primaryClass={cs.LG},
      url={https://arxiv.org/abs/2310.01557}, 
}

@misc{ni2024mixevalderivingwisdomcrowd,
      title={MixEval: Deriving Wisdom of the Crowd from LLM Benchmark Mixtures}, 
      author={Jinjie Ni and Fuzhao Xue and Xiang Yue and Yuntian Deng and Mahir Shah and Kabir Jain and Graham Neubig and Yang You},
      year={2024},
      eprint={2406.06565},
      archivePrefix={arXiv},
      primaryClass={cs.CL},
      url={https://arxiv.org/abs/2406.06565}, 
}
\end{document}